\documentclass[10pt,twocolumn,letterpaper]{article}

\usepackage{iccv}
\usepackage{times}
\usepackage{epsfig}
\usepackage{graphicx}
\usepackage{amsmath}
\usepackage{amssymb}
\usepackage[numbers]{natbib} 
\usepackage[inline, shortlabels]{enumitem} 
\usepackage{amsfonts,bm}

\usepackage{subcaption}

\usepackage[breaklinks=true,
            bookmarks=false,
            colorlinks = true,
            linkcolor = blue,
            urlcolor  = blue,
            citecolor = blue,
            anchorcolor = blue]{hyperref}
            
\usepackage{pifont}

\iccvfinalcopy 

\ificcvfinal\fi

\usepackage{cuted}
\usepackage{capt-of}
\usepackage{dsfont}

\begin{document}

\title{LatentCLR: A Contrastive Learning Approach for Unsupervised Discovery of Interpretable Directions}

\renewcommand{\thefootnote}{\fnsymbol{footnote}}

\author{\stepcounter{footnote}\vspace{1mm}Oğuz Kaan Yüksel  $^{1,}$\thanks{Equal contribution. Author ordering determined by a coin flip.}
\hspace{0.75cm}
Enis Simsar $^{2,3,}$\footnotemark[2] 
\hspace{0.75cm}
Ezgi Gülperi Er $^{3}$
\hspace{0.75cm}
Pinar Yanardag $^{3}$
\\
$^1$EPFL
\hspace{2em} 
$^2$Technical University of Munich
\hspace{2em} 
$^3$Boğaziçi University
\\
{\tt\small oguz.yuksel@epfl.ch}
\hspace{0.5em}
{\tt\small enis.simsar@tum.de}
\hspace{0.5em}
{\tt\small ezgi.er@boun.edu.tr}
\hspace{0.5em}
{\tt\small yanardag.pinar@gmail.com}
\vspace{-2cm}
}

\maketitle

\ificcvfinal\thispagestyle{empty}\fi
\vspace*{-\baselineskip}
\begin{strip}\centering
 \begin{minipage}{.46\textwidth}
        \centering
        \includegraphics[width=\textwidth]{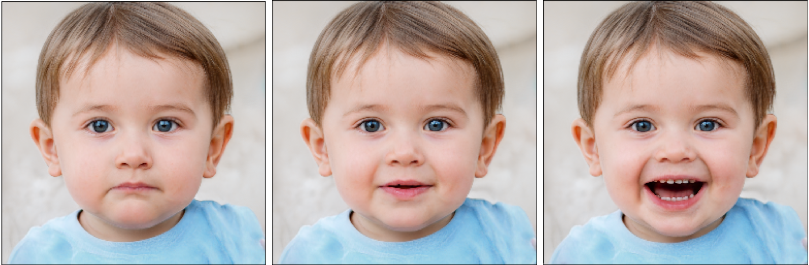}
        \vspace{-0.75cm}\captionof*{figure}{\small{[StyleGAN2] Smile on FFHQ} }
\end{minipage}
\begin{minipage}{0.46\textwidth}
        \centering
        \includegraphics[width=\textwidth]{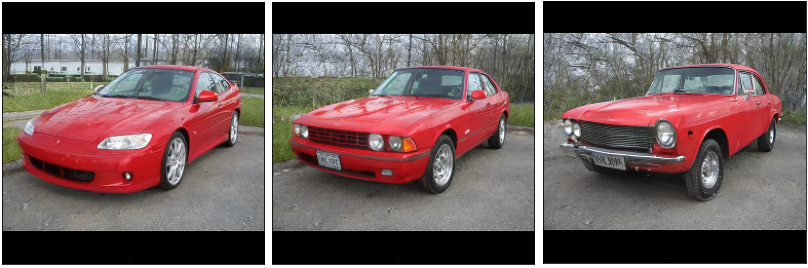}
        \vspace{-0.75cm}\captionof*{figure}{\small{[StyleGAN2] Car type on LSUN Cars}}
\end{minipage}
\begin{minipage}{.46\textwidth}
        \centering
        \includegraphics[width=\textwidth]{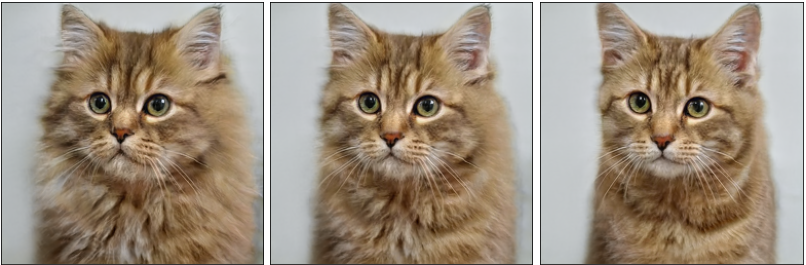}
        \vspace{-0.75cm}\captionof*{figure}{\small{[StyleGAN2] Fluffiness on LSUN Cats}}
\end{minipage}
\begin{minipage}{0.46\textwidth}
        \centering
        \includegraphics[width=\textwidth]{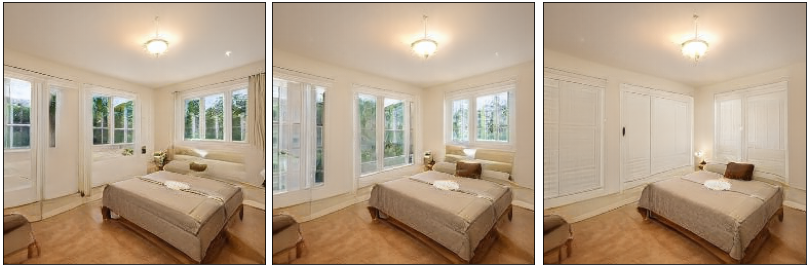}
        \vspace{-0.75cm}\captionof*{figure}{\small{[StyleGAN2] Window on LSUN Bedrooms}}
\end{minipage}
\begin{minipage}{.46\textwidth}
        \centering
        \includegraphics[width=\textwidth]{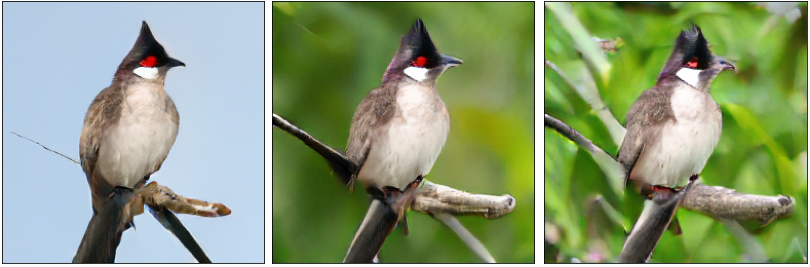}
        \vspace{-0.75cm}\captionof*{figure}{\small{[BigGAN] Background removal on ImageNet \textit{Bulbul}}}
\end{minipage}
    \begin{minipage}{0.46\textwidth}
        \centering
        \includegraphics[width=\textwidth]{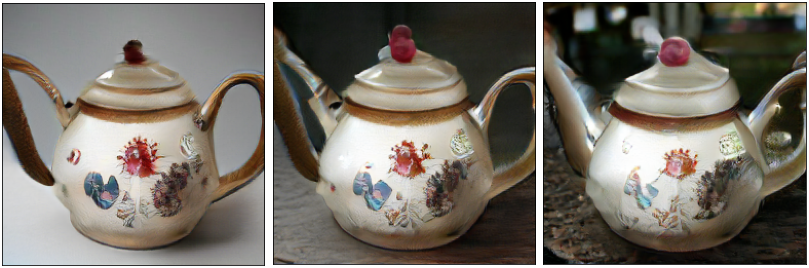}
        \vspace{-0.75cm}\captionof*{figure}{\small{[BigGAN] Background removal transferred from \textit{Bulbul}}}
\end{minipage}
\captionof{figure}{Interpretable directions discovered in StyleGAN2 \cite{karras2020analyzing} and BigGAN \cite{BigGAN}. Left and right images of each triplet are obtained by moving the latent code of the image, shown in the middle, towards negative and positive directions, respectively. Our directions are transferable, such as background removal learned on \textit{Bulbul} class.
\label{fig:teaser}}
\end{strip}

\vspace{-0.5cm}
\begin{abstract} 
\vspace{-0.3cm}
Recent research has shown that it is possible to find interpretable directions in the latent spaces of pre-trained Generative Adversarial Networks (GANs). These directions enable controllable image generation and support a wide range of semantic editing operations, such as zoom or rotation. The discovery of such directions is often done in a supervised or semi-supervised manner and requires manual annotations which limits their use in practice. In comparison, unsupervised discovery allows finding subtle directions that are difficult to detect a priori. In this work, we propose a contrastive learning-based approach to discover semantic directions in the latent space of pre-trained GANs in a self-supervised manner. Our approach finds semantically meaningful dimensions comparable with state-of-the-art methods.
\end{abstract}
 
\vspace{-0.5cm}
\section{Introduction}
Generative Adversarial Networks (GANs) \cite{NIPS2014_5423} are powerful image synthesis models that have revolutionized generative modeling in computer vision. Due to their success in synthesizing high-quality images, they are widely used for various visual tasks, including image generation \cite{DBLP:journals/corr/abs-1710-10916}, image manipulation \cite{wang2017highresolution},  de-noising \cite{wang2019spatial,li2019single}, upscaling image resolution \cite{Sun_2020}, and domain translation \cite{CycleGAN}.

Until recently, GAN models have generally been interpreted as black-box models, without the ability to control the generation of images. Some degree of control can be achieved by training conditional models such as \cite{mirza2014conditional} and changing conditions in generation-time. Another approach is to design models that generate a more disentangled latent space such as in InfoGAN \cite{chen2016infogan} where each latent dimension controls a particular attribute. However, these approaches require labels and provide only limited control, depending on the granularity of available supervised information. 

Albeit some progress has been done, the question of what knowledge GANs learn in the latent representation and how these representations can be used to manipulate images is still an ongoing research question. Early attempts to explicitly control the underlying generation process of GANs include simple approaches, such as modifying the latent code of images \cite{radford2015unsupervised} or interpolating latent vectors \cite{StyleGAN}. Recently, several approaches have been proposed to explore the structure of latent space in GANs in a more principled way \cite{jahanian2019steerability,shen2020interfacegan,harkonen2020ganspace, voynov2020unsupervised,plumerault2020controlling}. Most of these works discover domain-agnostic interpretable directions such as \textit{zoom, rotation}, or \textit{translation}, while other find domain-specific directions such as changing \textit{gender, age} or \textit{expression} on facial images. Typically, such methods either identify or optimize for directions and then shift the latent code in these directions to increase or decrease target semantics in the image.

In this paper, we introduce \textbf{LatentCLR}, an optimization-based approach that uses a self-supervised contrastive objective to find interpretable directions in GANs. In particular, we use the \emph{differences} caused by an edit operation on the feature activations to optimize the identifiability of each direction. Our contributions are as follows:

\begin{itemize} 
    \item We propose to use contrastive learning on feature divergences to discover interpretable directions in the latent space of pre-trained GAN models such as StyleGAN2 and BigGAN. 
    \item We show that our method can find distinct and fine-grained directions on a variety of datasets, and that the obtained directions are highly transferable between ImageNet \cite{ImageNet} classes. 
    \item We make our implementation publicly available to encourage further research in this area: \url{https://github.com/catlab-team/latentclr}.
\end{itemize}

The rest of this paper is organized as follows. Section \ref{sec:related_work} discusses related work. Section \ref{sec:methodology} introduces our contrastive framework. Section \ref{sec:experiments} presents our quantitative and qualitative results. Section \ref{sec:limitations} discusses the limitations of our work and Section \ref{sec:conclusion} concludes the paper.

\section{Related Work}
\label{sec:related_work}
In this section, we introduce generative adversarial networks and discuss latent space manipulation methods.

\subsection{Generative Adversarial Networks} 
Generative Adversarial Networks (GANs) consist of a generator and a discriminator for mapping the real world to the generative space \cite{NIPS2014_5423}. The discriminator part of the network tries to detect whether images are from the training dataset or synthetic, while the generative part tries to generate images that are similar to the dataset. StyleGAN \cite{StyleGAN} and StyleGAN2 are among the popular GAN approaches that are capable of generating high-quality images. They use a mapping network consisting of an 8-layer perceptron that aims to map the input latent code to an intermediate latent space. Another popular GAN model is BigGAN \cite{BigGAN}, a large-scale model trained on ImageNet. Similar to StyleGAN2, it also makes use of intermediate layers by using the latent vector as input, also called \textit{skip-z} inputs, as well as a class vector. Due to its conditional architecture, it can generate images in a variety of categories from ImageNet. In this paper, we work with pre-trained StyleGAN2 and BigGAN models.

\begin{figure*}[t!]
\centering
        \includegraphics[width=1\textwidth]{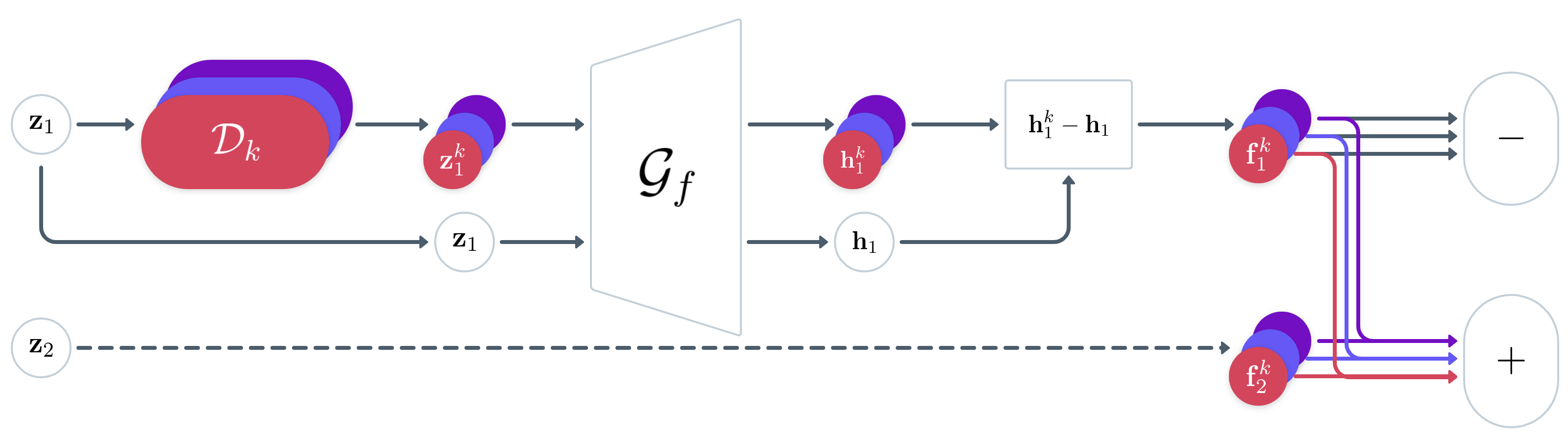}
 \caption{\textbf{Illustration of LatentCLR.} First, each latent code is passed through direction models (denoted with $\mathcal{D}_k$) and up to a target feature layer of the GAN (denoted with $\mathcal{G}_f$) to obtain intermediate representations of edited codes. Then, the effects of direction models are computed by subtracting the representation of the original latent code. Finally, pairs produced by the same model are considered as positive, and others as negative, in a contrastive loss.}
\label{fig:framework}
\end{figure*}

\subsection{Latent Space Navigation}

Recently, several strategies have been proposed to manipulate the latent structure of pre-trained GANs. These methods manipulate images in different ways by editing the latent code and can be divided into two groups.

\paragraph{Supervised Setting.}
Supervised approaches typically use pre-trained classifiers to guide optimization-based learning to discover interpretable directions that specifically manipulate the properties of interest. InterfaceGAN \cite{shen2020interfacegan} is a supervised approach that benefits from labeled data including \textit{gender, facial expression} and \textit{age}. It trains binary Support Vector Machines (SVM) \cite{noble2006support} on each label and interprets the normal vectors of the obtained hyperplanes as latent directions. GANalyze \cite{goetschalckx2019ganalyze} finds directions for cognitive image properties for a pre-trained BigGAN model using an externally trained \textit{assessor} function. Feedback from the assessor guides the optimization process, and the resulting optimal direction allows manipulation of the desired cognitive attributes. StyleFlow \cite{abdal2021styleflow} uses attribute-conditioned continuous normalizing flows that use labels to find edit directions in the latent space of GANs. 

\paragraph{Unsupervised Setting.} One of the unsupervised works proposed by \cite{voynov2020unsupervised}  discovers meaningful directions using a classifier-based approach. Given a particular manipulation, the classifier tries to detect which particular direction is applied. At the end of the optimization process, the method learns disentangled directions. Ganspace \cite{harkonen2020ganspace} is a sampling based unsupervised method where latent vectors are randomly selected from the intermediate layers of BigGAN and StyleGAN models. Then they propose to use Principal Component Analysis (PCA) \cite{wold1987principal} to find principal components that are interpreted as semantically meaningful directions. The principal components lead to a variety of useful manipulations, including \textit{zoom, rotation} in BigGAN or changing \textit{gender, hair color, or age} in StyleGAN models. SeFa \cite{shen2020closed} follows a related approach, using a closed-form solution that specifically optimizes the intermediate weight matrix of the pre-trained GAN model. They obtain interpretable directions in the latent space by computing the eigenvectors of the first projection matrix and selecting the eigenvectors with the largest eigenvalues. A different closed-form solution is proposed by \cite{spingarn2020gan} that discovers directions without optimization. Another work proposed by \cite{jahanian2019steerability} exploits task-specific edit functions. They start by applying an editing operation to the original image, e.g., zoom, and minimize the distance between the original image and the edited image to learn a direction that leads to the desired editing operation. \cite{tov2021designing} provides directions by using a GAN inversion model and aims to change the image in a particular direction without changing the remaining properties. Instead of working on latent codes, \cite{cherepkov2021navigating} uses the space of generator parameters to discover semantically meaningful directions. \cite{nitzan2020face} uses a fixed decoder and  trains an encoder to decouple the processes of disentanglement and synthesis. \cite{khrulkov2020disentangled} uses
post-hoc disentanglement that requires little to no hyperparameters. \cite{lin2020infogan} is a variant of InfoGAN that uses a contrastive regularizer and aims to make the elements of the latent code set clearly identifiable from one another. A concurrent work to ours is proposed by \cite{ren2021generative} which uses  an entropy-based domination loss and a hard negatives flipping strategy to achieve disentanglement.

\section{Methodology}
\label{sec:methodology}

In this section, we first introduce preliminaries of contrastive learning and then discuss details of our method.

\subsection{Contrastive Learning}
Contrastive learning has recently become popular due to leading state-of-the-art results in various unsupervised representation learning tasks. It aims to learn representations by contrasting positive pairs against negative pairs \cite{hadsell2006dimensionality} and is used in various computer vision tasks, including data augmentation \cite{chen2020simple,oord2018representation}, or diverse scene generation \cite{tian2019contrastive}. The core idea of contrastive learning is to move the representations of similar pairs near and dissimilar pairs far. 

In this work, we follow a similar approach to SimCLR framework \cite{chen2020simple} for contrastive learning. SimCLR consists of four main components: a stochastic data augmentation method that generates positive pairs $(\mathbf{x}, \mathbf{x}^+)$, an encoding network $f$ that extracts representation vectors out of augmented samples, a small projector head $g$ that maps representations to the loss space, and a contrastive loss function $\ell$ that enforces the separation between positive and negative pairs. Given a random mini-batch of $N$ samples, SimCLR generates $N$ positive pairs using the specified data augmentation method. For all positive pairs, the remaining $2(N-1)$ augmented samples are treated as negative examples. Let $\mathbf{h}_i = f(\mathbf{x}_i)$ be the representations of all $2N$ samples and $\mathbf{z}_i = g(\mathbf{h}_i)$ be the projections of these representations. Then, SimCLR considers the average of the \textit{NT-Xent} loss \cite{sohn2016improved, oord2018representation} over all positive pairs $(\mathbf{x}_i, \mathbf{x}_j)$:

\begin{equation}
    \ell(\mathbf{x}_{i}, \mathbf{x}_j) = -\log \frac{\exp(\text{sim}(\mathbf{z}_i, \mathbf{z}_j)/\tau)}{\sum_{k=1}^{2N}\mathds{1}_{[k\neq i]}\exp({\text{sim}(\mathbf{z}_i,\mathbf{z}_k)/\tau})}
\label{eq:ntxloss}
\end{equation}

where $\text{sim}(\mathbf{u}, \mathbf{v}) = \mathbf{u}^T\mathbf{v}/\|\mathbf{u}\|\|\mathbf{v}\|$ is the cosine similarity function, $\mathds{1}_{[k\neq i]} \in \{0,1\}$ is an indicator function that takes the value 1 only when $k=i$, and $\tau$ is the temperature parameter. The two networks $f$ and $g$ are trained together. Intuitively, $g$  learns a mapping to a space where cosine similarity represents semantic similarity and NT-Xent objective encourages \textit{identifiability} of positive pairs among all other negative examples. This, in turn,  forces $f$ to learn representations that are invariant to the given data augmentations, up to a nonlinear mapping and cosine similarity.

\subsection{Latent Contrastive Learning (LatentCLR)}

Consider a pre-trained GAN, expressed as a mapping function $\mathcal{G}: \mathcal{Z} \to \mathcal{X}$ where $\mathcal{Z}$ is the latent space, usually associated with a prior distribution such as the multivariate Gaussian distribution, and $\mathcal{X}$ is the target image domain. Given a latent code $\mathbf{z}$ and its generated image $\mathbf{x} = \mathcal{G}(\mathbf{z})$, we look for edit directions $\Delta \mathbf{z}$ such that the image $\mathbf{x}' = \mathcal{G}(\mathbf{z}+\Delta\mathbf{z})$ has semantically meaningful changes with respect to $\mathbf{x}$ while preserving the identity of $\mathbf{x}$. Similar to \cite{harkonen2020ganspace, shen2020closed, voynov2020unsupervised}, we limit ourselves to the unsupervised setting, where we aim to identify such edit directions without external supervision, as in \cite{goetschalckx2019ganalyze, shen2020interfacegan, jahanian2019steerability}. 

Our intuition is to optimize an \emph{identifiability-based} heuristic, similar to \cite{voynov2020unsupervised}, in an intermediate representation space of a pre-trained GAN to find a diverse set of interpretable directions. We search for edit directions $\Delta\mathbf{z}_1, \cdots, \Delta\mathbf{z}_K$, $K > 1$, that have distinguishable effects in the target representation layer. For this end, we calculate \emph{differences} in representations induced by each direction and use a contrastive-learning objective to maximize the identifiability of directions.

More specifically, we generalize \textit{directions} with potentially more expressive conditional mappings called \textit{direction models}. Then, our final approach consist of \begin{enumerate*}[series = tobecont, itemjoin = \quad, label=(\roman*)]
\item concurrent directions models that apply edits to the given latent codes,
\item a target feature layer $f$ of a pre-trained GAN $\mathcal{G}$ that will be used to evaluate direction models, and as well as 
\item a contrastive learning objective for measuring identifiability of each direction model
\end{enumerate*}.
See Figure~\ref{fig:framework} for a high-level visualization.

\paragraph{Direction models.} The direction model is a mapping $\mathcal{D}: \mathcal{Z} \times \mathbb{R} \to \mathcal{Z}$ that takes latent codes along with a desired edit \textit{magnitude} and outputs edited latent codes, i.e. $\mathcal{D}: (\mathbf{z}, \alpha) \to \mathbf{z}+\Delta\mathbf{z}$, where $\|\Delta\mathbf{z}\| \propto \alpha$. We consider three alternative methods for choosing the direction model: \textit{global, linear} and \textit{non-linear}, which are defined as follows:
\begin{itemize}
    \item \emph{Global.} We learn a fixed direction $\mathbf{\theta}$ independent of the latent code $\mathbf{z}$.
    $$\mathcal{D}(\mathbf{z}, \alpha) = \mathbf{z} + \alpha \frac{\mathbf{\theta}}{\|\mathbf{\theta}\|}$$

    \item \emph{Linear.} We learn a matrix $\mathbf{M}$ to output a conditional direction on the latent code $\mathbf{z}$, which is a linear dependency.
    $$\mathcal{D}(\mathbf{z}, \alpha) = \mathbf{z} + \alpha \frac{\mathbf{M}\mathbf{z}}{\|\mathbf{M}\mathbf{z}\|}$$
    
    \item \emph{Nonlinear.} We learn a multi-layer perceptron, represented by $\mathbf{NN}$, to represent an arbitrarily complex dependency between direction and latent code.  
    
    $$\mathcal{D}(\mathbf{z}, \alpha) = \mathbf{z} + \alpha \frac{\mathbf{NN}(\mathbf{z})}{\|\mathbf{NN}(\mathbf{z})\|}$$
    
\end{itemize}

\begin{figure*}[ht]
\centering
\begin{minipage}{0.49\textwidth}
        \includegraphics[width=\textwidth]{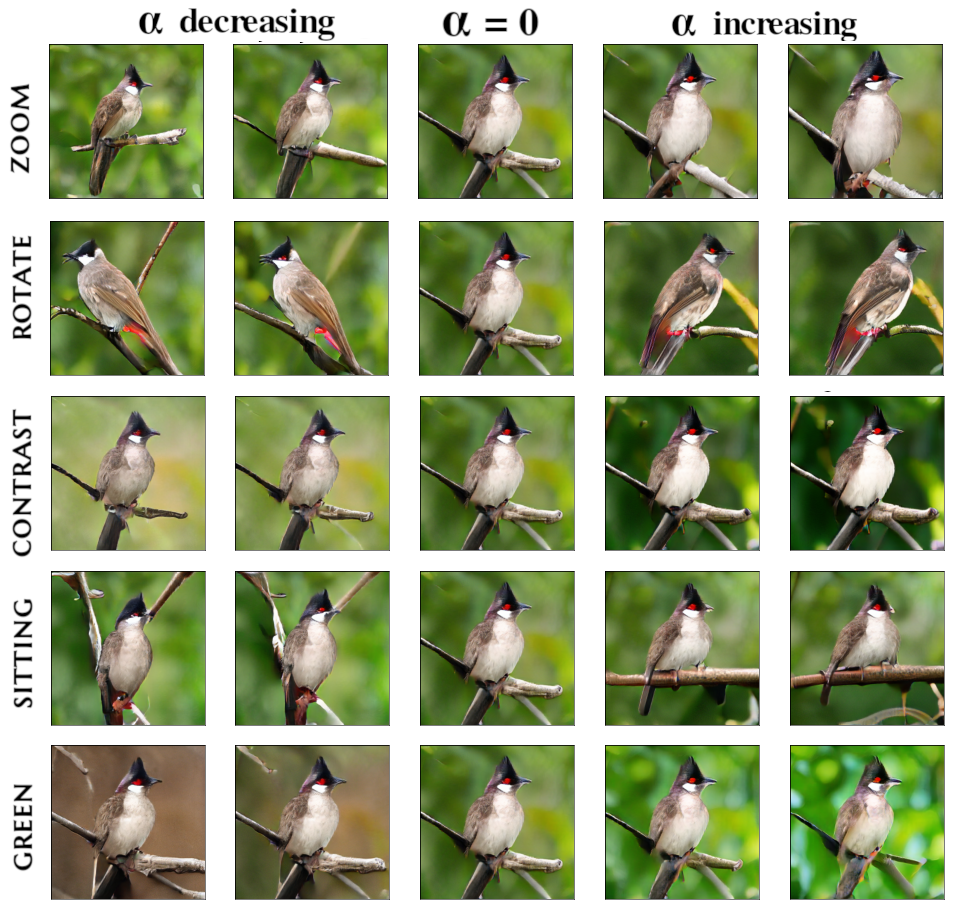}
      \vspace{-0.75cm} \caption*{(a)}
\end{minipage}
\begin{minipage}{.49\textwidth}
        \centering
        \includegraphics[width=\textwidth]{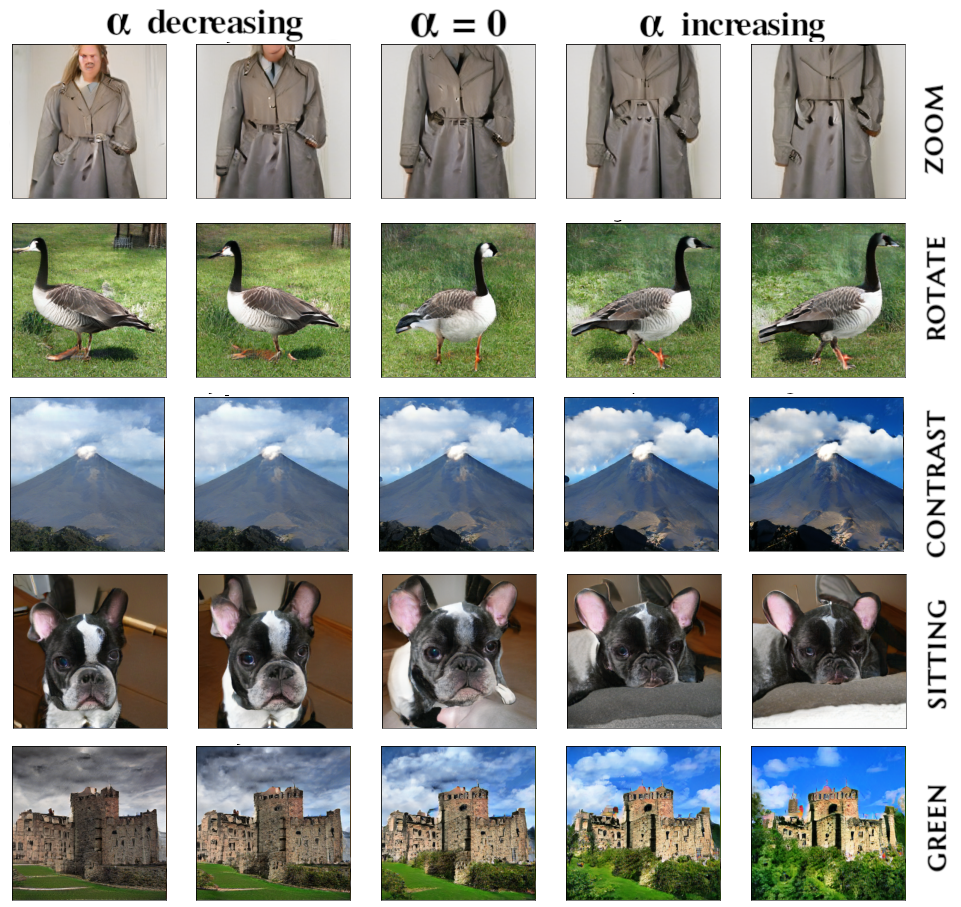}
      \vspace{-0.75cm} \caption*{(b)}
\end{minipage}%
\caption{(a) Directions for general image editing operations such as \textit{zoom}, or \textit{rotation} discovered from the ImageNet Bulbul class, where we shift the latent code in a particular direction with increasing or decreasing $\alpha$. (b) Transferred directions from the \textit{Bulbul} class to various other ImageNet classes.}
\label{fig:biggan_bulbul}
\end{figure*}
Note that for all options, we apply $\ell_2$ normalization and thus the magnitudes given by $\alpha$ correspond to the direct $\ell_2$ distances from the latent code.
 
The first option, \textit{Global}, is in principle the most limited since it can only find fixed directions and thus edits images without considering the latent code $\mathbf{z}$. However, we note that it is still able to capture common directions such as \textit{zoom, rotation}, or \textit{background removal} on BigGAN. The second option, \textit{Linear} is able to generate \textit{conditional} directions given latent code $\mathbf{z}$, but is still limited for capturing finer-grained directions. The third option is an extension of the \textit{Linear} direction model, where we use a neural network that models the dependency between the direction and the given latent code. \\

\noindent\textbf{Target feature differences.} For each latent code $\mathbf{z}_i, 1 \leq i \leq N$ in the mini-batch of size $N$, we compute $K$ distinct edited latent codes:
$$\mathbf{z}_i^k = \mathcal{D}(\mathbf{z}_i, \alpha).$$
Then, we calculate corresponding intermediate feature representations,
$$\mathbf{h}_i^k = \mathcal{G}_f(\mathbf{z}_i^k),$$
where $\mathcal{G}_f$ is feed-forward of GAN up to the target layer $f$.
Next, we compute the feature divergences w.r.t. the original latent code,
$$\mathbf{f}_i^k = \mathbf{h}_i^k - \mathcal{G}_f(\mathbf{z_i}).$$

\noindent\textbf{Objective function.} 
For each edited latent code $\mathbf{z}_i^k$, we define the following loss:
 
\begin{equation*}
    \ell(\mathbf{z}_i^k) = -\log\frac{\sum_{j=1}^{N}\mathds{1}_{[j\neq i]} \exp\big(\text{sim}(\mathbf{f}_i^k, \mathbf{f}_j^k)/\tau\big) }{\sum_{j=1}^{N}\sum_{l=1}^{K}\mathds{1}_{[l\neq k]} \exp\big(\text{sim}(\mathbf{f}_i^k, \mathbf{f}_j^l)/\tau\big) }
\end{equation*}

The intuition behind our objective function is as follows. All feature divergences obtained with the same latent edit operation $1 \leq k \leq K$, i.e., each of $\mathbf{f}_1^k, \mathbf{f}_2^k, \cdots \mathbf{f}_N^k$, are considered as \textit{positive} pairs and contribute to the numerator. All other pairs obtained with a different edit operation, e.g., $\mathbf{f}_1^k \neq \mathbf{f}_N^l$, $l \neq k$, are considered as \textit{negative} pairs and contribute to the denominator. This can be viewed as a generalization of the NT-Xent loss (Eq. \ref{eq:ntxloss}), where we have $N$-tuples of groups, one for each direction model. With this generalized contrastive loss, we enforce latent edit operations to have orthogonal effects on the features.\\

\noindent\textbf{Utilizing Layer-wise Styles.}
Ganspace \cite{harkonen2020ganspace} discovered that the layer-wise structure of StyleGAN2 and BigGAN models can be used for fine-grained editing. By applying their directions only to a limited set of layers in test-time, they achieve less entanglement in editing and superior granularity. SeFa \cite{shen2020closed} can find more detailed directions by concatenating weight matrices and identifying eigenvectors. In contrast to Ganspace, our method can consider such layer-wise structure in training time. And, in contrast to SeFa, our method, additionally, can fuse the effects of all selected layers into the target feature layer due to its flexible optimization-based objective.

\begin{figure*}[t!]
\centering
\begin{minipage}{.46\textwidth}
        \centering
        \includegraphics[width=\textwidth]{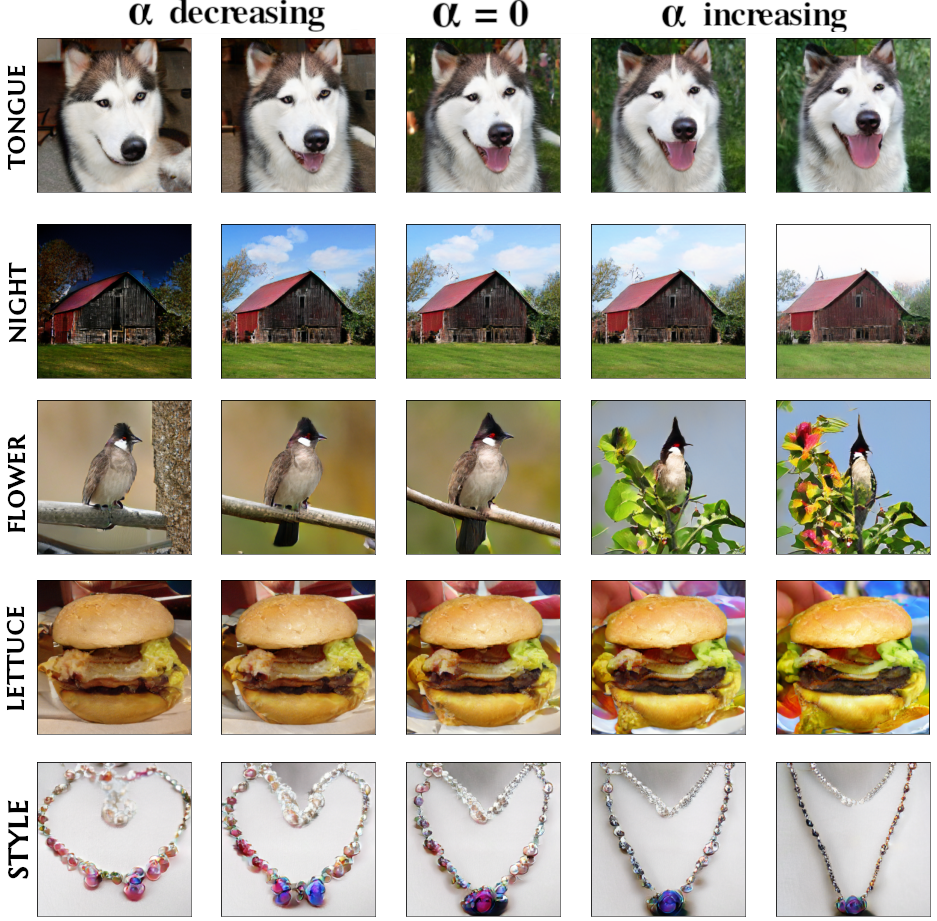}
      \vspace{-0.75cm} \caption*{(a)}
\end{minipage}%
\begin{minipage}{0.56\textwidth}
        \includegraphics[width=\textwidth]{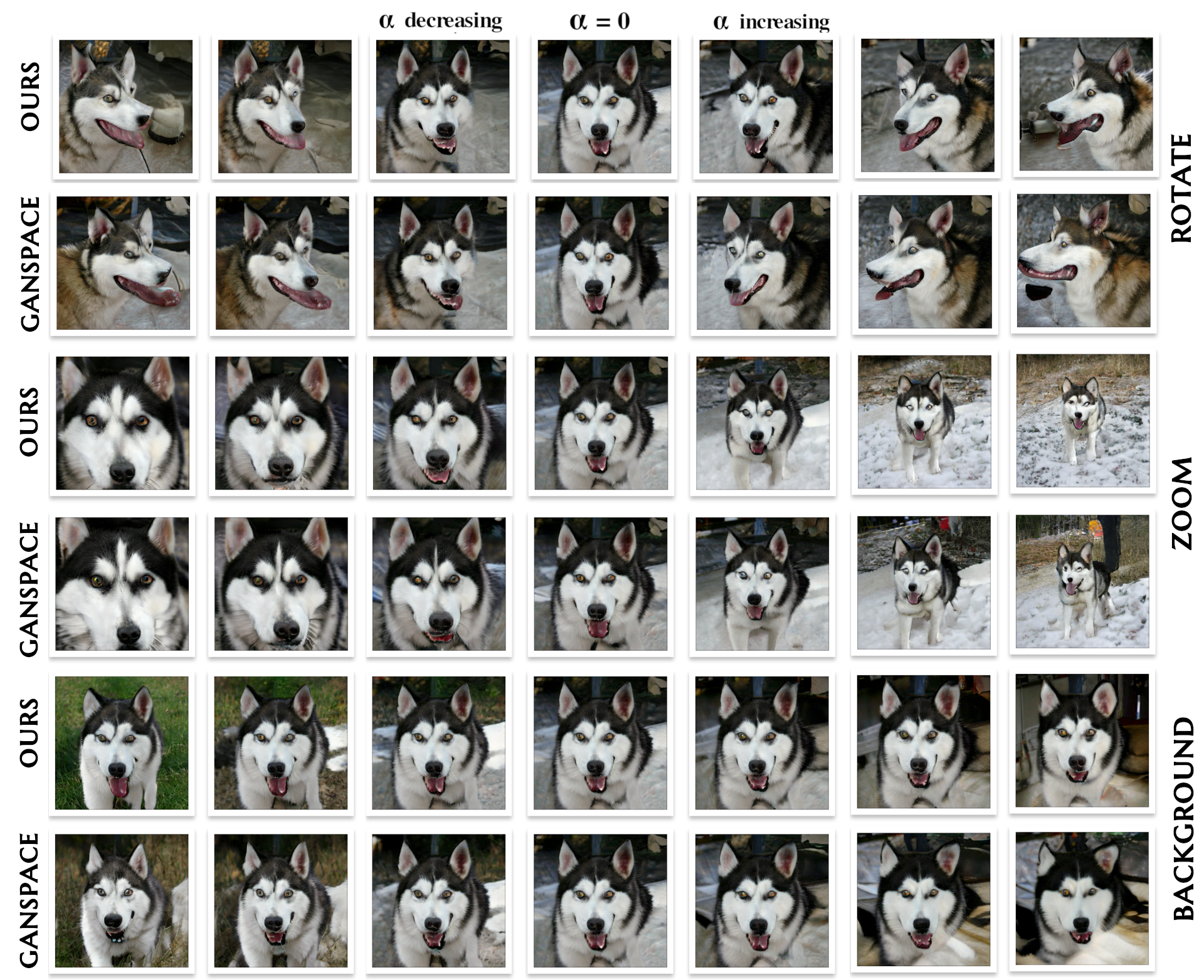}
      \vspace{-0.75cm} \caption*{(b)}
\end{minipage}
\caption{(a) Class-specific directions discovered by our method in several ImageNet classes on the BigGAN model. (b) A comparison of the directions \textit{rotate, zoom} and \textit{background change} between our method and Ganspace.}
\label{fig:biggan_ganspace}
\end{figure*}
\section{Experiments}
\label{sec:experiments}

We evaluate the proposed method for detecting semantically meaningful directions using different models and datasets. We apply the proposed model to BigGAN and StyleGAN2 on a wide range of datasets, including human faces (FFHQ) \cite{StyleGAN}, LSUN Cats, Cars, Bedrooms, Church and Horse datasets \cite{yu2015lsun} and ImageNet. 
We also compare our method with state-of-the-art unsupervised methods \cite{harkonen2020ganspace, shen2020closed} and conduct several qualitative and quantitative experiments to demonstrate the effectiveness of our approach. Next, we discuss our experimental setup and then present results on BigGAN and StyleGAN2 models.

\subsection{Experimental Setup}
For BigGAN experiments, we choose \textit{batch size} = 16, \textit{K} = 32 directions, \textit{output resolution} = 512, \textit{truncation} = 0.4, \textit{feature layer} = \textit{generator.layers.4} and train the models for 3 epochs (each epoch includes 100k iterations), which takes about 20 minutes. For StyleGAN2 experiments, we choose \textit{batch size} = 8, \textit{K} = 100 directions, \textit{truncation} = 0.7, \textit{feature layer} = \textit{conv1} and train StyleGAN2 models for 5 epochs (each epoch corresponds to 10k iterations), which takes about 12 minutes. We use 1-3 dense layers consisting of units corresponding to latent space (128, 256, or 512), with ReLU and batch normalizations. For our experiments, we use PyTorch framework and two NVIDIA Titan RTX GPUs.\\

\noindent\textbf{Choice of K.} To learn $K$ different directions, we use $K$ copies of the same direction model.
We observe that using too many directions leads to repetitive directions, a similar observation made by \cite{voynov2020unsupervised}. For BigGAN, we used $K=32$ directions since the latent space is $128$-dimensional and most interpretable directions such as \textit{zoom, rotation} or \textit{translation} can be obtained with a relatively small number of directions. For StyleGAN2, we used $K=100$ directions since the latent space is 512-dimensional.\\

\noindent\textbf{Layers.} To avoid any bias between the competing methods Ganspace and SeFa, we use the same set of StyleGAN2 layers that comes from Ganspace. We also note that slight differences in re-scoring analysis or visuals might be caused due to applying different magnitudes of change across methods, as they are not directly convertible between methods. To minimize this effect, we use the same sigma values in Ganspace as specified for each direction in their public repository, and $\{-3,+3\}$ for SeFa as provided in the official implementation.\\

\subsection{Results on BigGAN}
We evaluate our approach using a pre-trained BigGAN model conditionally trained on 1000 ImageNet classes. We trained our model on an arbitrary class, \textit{Bulbul}, and obtained $K=32$ directions.

\paragraph{Qualitative Results.} Our visual analysis shows that our model is able to detect several semantically meaningful directions such as \textit{zoom, rotation, contrast} as well as some finer-grained features such as \textit{background removal, sitting}, or \textit{green background} (see Figure \ref{fig:teaser} and Figure \ref{fig:biggan_bulbul} (a)). As can be seen from the figures, our method is able to manipulate the original image (labelled $\alpha$=0) by shifting the latent code towards the interpretable direction (increasing $\alpha$) or backwards (decreasing $\alpha$).

\paragraph{Transferability of directions.} After verifying that the directions obtained for the class \textit{Bulbul} are capable of manipulating the latent codes for multiple semantic directions, we investigated how transferable the discovered directions are to other ImageNet classes.

Our visual analysis shows that the directions learned from the \textit{Bulbul} class are applicable to a variety of ImageNet classes and are able to zoom a \textit{Trenchcoat}, rotate a \textit{Goose}, apply contrast to \textit{Volcano}, add greenness to \textit{Castle} classes (see Figure \ref{fig:biggan_bulbul} (b)) as well as removing the background from a \textit{Teapot} object (see Figure \ref{fig:teaser}).  An interesting transferred direction is \textit{Sitting} (see Figure \ref{fig:biggan_bulbul} (a)), which manipulates the latent code so that the bird in the Bulbul class sits on a tree branch when we increase $\alpha$. We find that this direction, when applied to the class \textit{Bulldog}, also causes the dog to stand up or sit down depending on how we move in the positive or negative direction (see Figure \ref{fig:biggan_bulbul} (b)).

\begin{figure*}[t!]
\centering
\begin{minipage}{.65\textwidth}
        \centering
        \includegraphics[width=\textwidth]{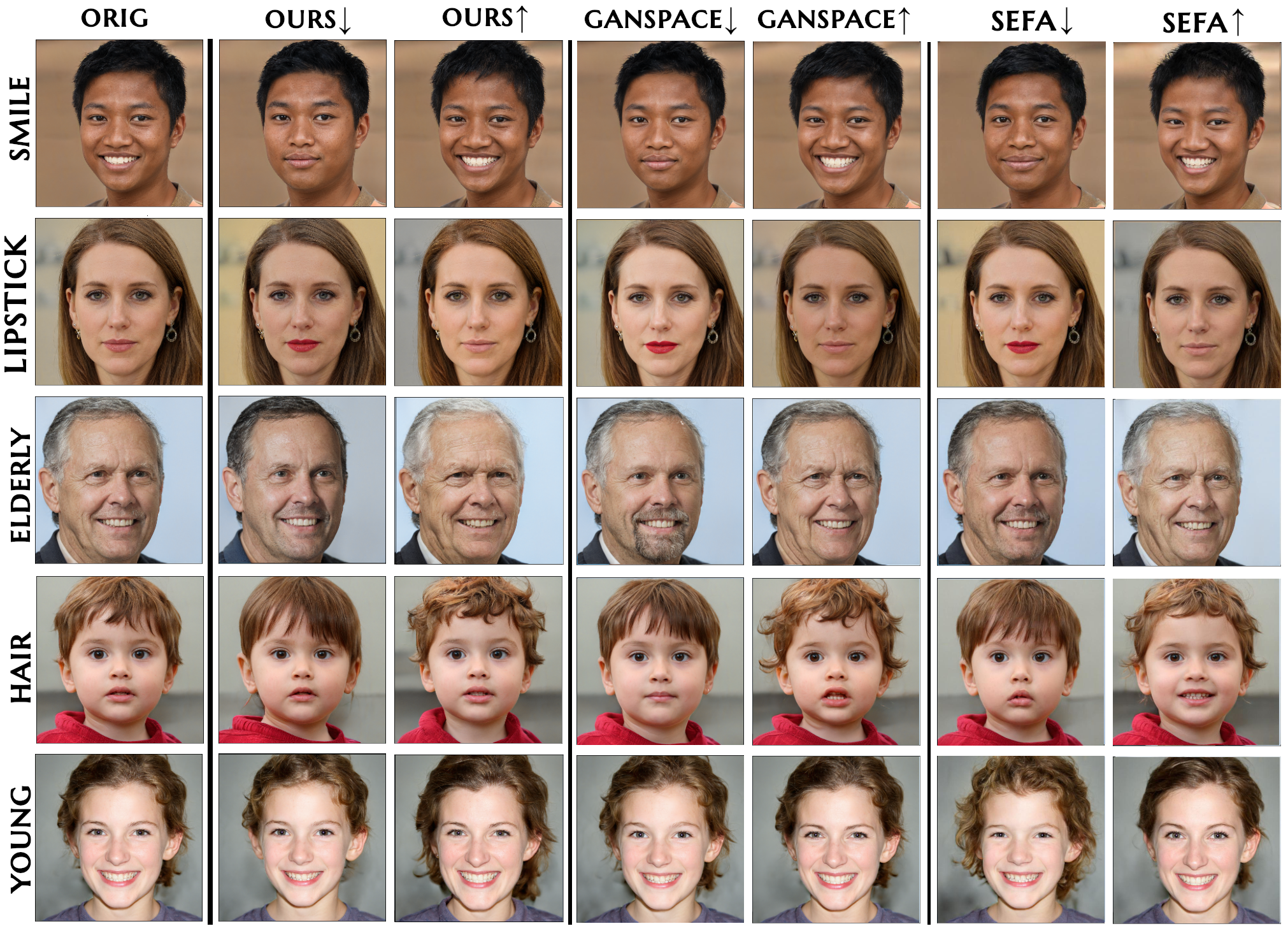}
      \vspace{-0.75cm} \caption*{(a)}
\end{minipage}%
\begin{minipage}{0.31\textwidth}
        \includegraphics[width=\textwidth]{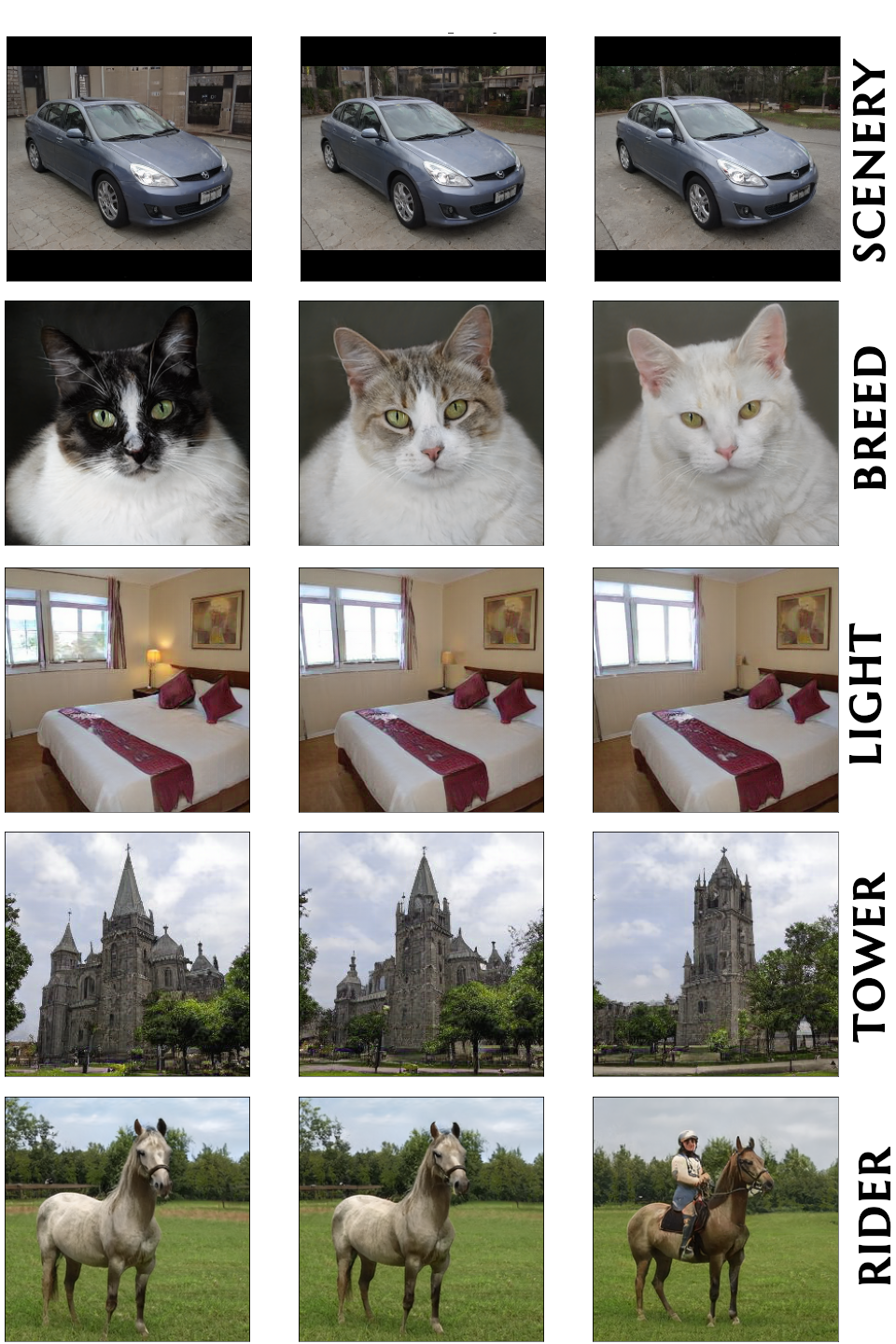}
      \vspace{-0.75cm} \caption*{(b)}
\end{minipage}
\caption{(a) Comparison of manipulation results on FFHQ dataset with Ganspace and SeFa methods. The leftmost image represents the original image, while images denoted with $\uparrow$ and $\downarrow$ represent the edited image moved in the positive or negative direction. (b) Directions discovered by our method in various LSUN \cite{yu2015lsun} datasets. }
\label{fig:stylegan_main}
\end{figure*}
\paragraph{Diversity of the directions.} Next, we investigate whether we can discover unique directions when we train our model on different ImageNet classes. Figure \ref{fig:biggan_ganspace} (a) shows some directions discovered by our model, such as adding \textit{tongue} on \textit{Husky} class, changing the \textit{time of day} in \textit{Barn} class, adding \textit{flowers} in \textit{Bulbul} class, adding \textit{lettuce} on \textit{Cheeseburger} class, or changing the \textit{shape}  in \textit{Necklace} class.

\begin{figure}[t!]
\centering
        \includegraphics[width=1\columnwidth]{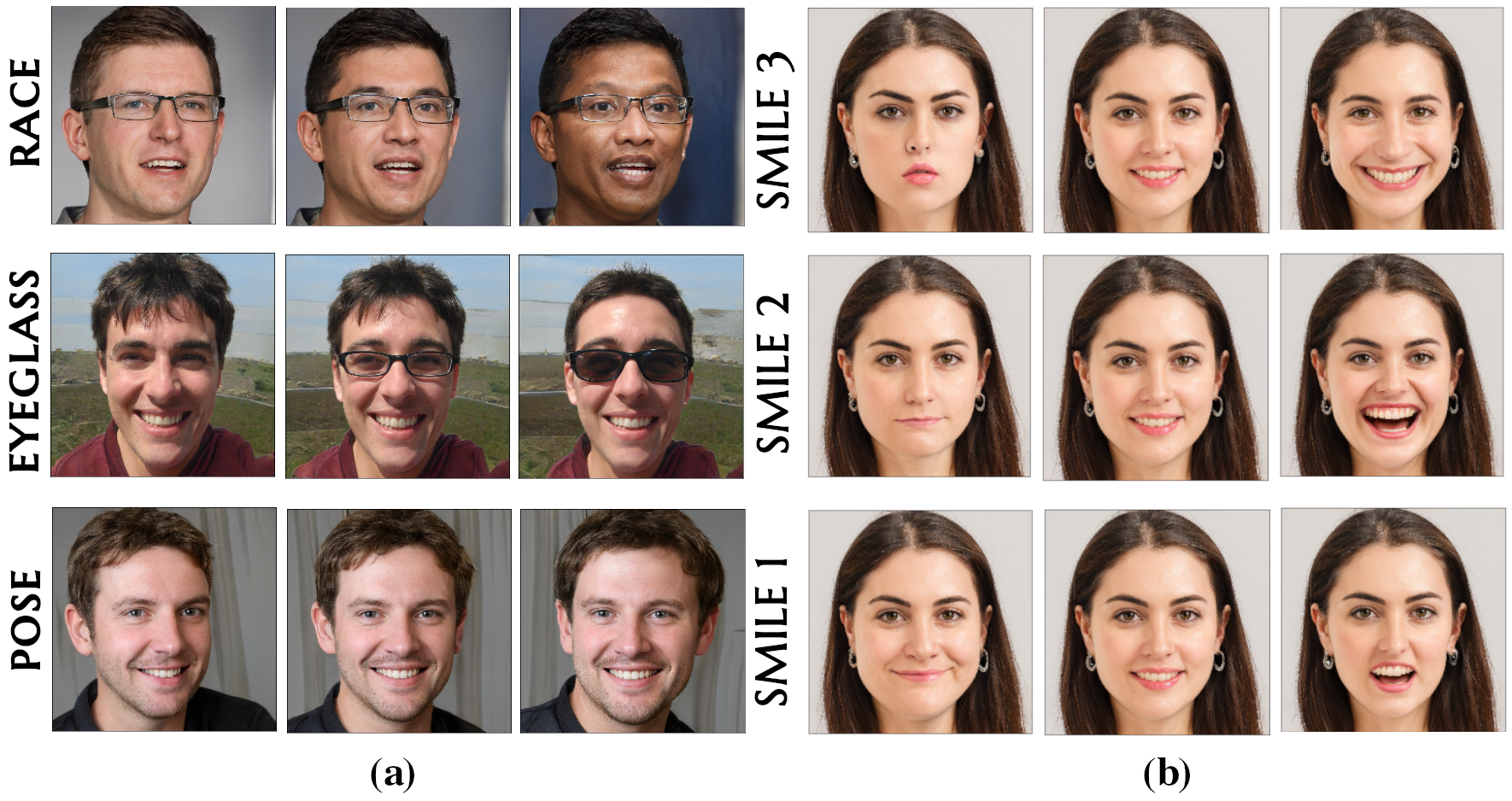}
 \caption{Additional directions discovered by our method in FFHQ dataset with StyleGAN2 (left). Different types of \textit{smile} directions (denoted with \textit{SMILE 1-3}) discovered by the non-linear approach (right). }
\label{fig:stylegan_small}
\end{figure}

\paragraph{Comparison with other methods.} We visually compare the directions obtained with our method to Ganspace using the  \textit{Husky} class (see Figure \ref{fig:biggan_ganspace} (b)). For each direction in Ganspace, we use the parameters given in the open-source implementation\footnote{\url{https://github.com/harskish/ganspace}}. For both methods, we use the same initial image (represented by $\alpha=0$) and move the latent code towards the direction (as $\alpha$ increases) and far away from the direction (as $\alpha$ decreases). We used the original $\alpha$ settings (i.e., the \textit{sigma} setting) for each direction as provided in the Ganspace implementation to avoid any bias that may be caused by tuning the parameter. We find that both methods achieve similar manipulations for the \textit{zoom, rotation}, and \textit{background change} directions, while our method causes less entanglement. For example, we note that Ganspace tends to increase \textit{tongue} with \textit{rotation} or add background objects with increasing \textit{zoom}.

\paragraph{User Study.} To understand how the directions found by our method match human perception, we conduct a user study on Amazon's Mechanical Turk platform where each participant is shown a randomly selected $10$ images out of a set of $100$ randomly generated images on BigGAN's \textit{Bulbul} class. Participants are shown the original image in the center and $-\alpha$ and $+\alpha$ values on the left and right sides, respectively. 
Following the same approach as \cite{shen2020closed}, we ask $n=100$ users the following questions: \textit{"Question 1: Do you think there is an obvious content change on the left and right images comparing to the one in the middle?"} and \textit{"Question 2: Do you think the change on the left and right images comparing to the one in the middle is semantically meaningful?"}. Each question is associated with \textit{Yes/Maybe/No} options, and the order of the questions is also randomized. 
Our user study shows that for \textit{Question 1}, participants answered $17.43$ on average with "Yes", $7.43$ with "Maybe," and $5.75$ with "No". For \textit{Question 2}  we got \textit{Yes} on average $14.37$ directions, \textit{Maybe} on average $10.03$ directions, and \textit{No} on average $6.21$ directions. These results indicate that out of $K=32$ directions, participants found $82\%$ semantically meaningful and $80.59\%$ directions  containing an obvious content change.

\begin{table}[ht!]
\begin{center}
\begin{tabular}{|l|c|c|c|c|c|c|}
\hline
\textbf{Model} & \textbf{Ganspace} & \textbf{SeFa} & \textbf{Ours}  \\
\hline\hline
$\uparrow$ Smile    & 0.99 $\pm 0.11$ & 0.89 $\pm 0.31$ & 0.99 $\pm 0.11$ \\
$\uparrow$ Age       & 0.32 $\pm 0.12$ & 0.38 $\pm 0.15$ & 0.42 $\pm 0.13$ \\
$\uparrow$ Lipstick    & 0.58 $\pm 0.49$ & 0.55 $\pm 0.49$ & 0.66 $\pm 0.47$ \\
\hline
\hline
$\downarrow$  Smile & 0.05 $\pm 0.21$ & 0.50 $\pm 0.50$ & 0.11 $\pm 0.32$ \\
$\downarrow$ Age     & 0.23 $\pm 0.08$ & 0.23 $\pm 0.06$ & 0.23 $\pm 0.07$ \\
$\downarrow$ Lipstick    & 0.35 $\pm 0.47$ & 0.35 $\pm 0.48$ & 0.36 $\pm 0.48$ \\
\hline
\end{tabular}
\end{center}
\caption{ Re-scoring Analysis on Ganspace, SeFA and our method where we compare \textit{Smile, Age} and \textit{Lipstick} attributes using FFHQ dataset on StyleGAN2.}
\label{tab:stylegan2}
\end{table}
\subsection{Results on StyleGAN2}
 
We apply our method on StyleGAN2 to a wide range of datasets and compare our results with state-of-the-art unsupervised methods Ganspace and SeFa.

\paragraph{Qualitative Results.} First, we visually examined the directions found by our method in several datasets, including FFHQ, LSUN Cars, Cats, Bedroom, Church, and Horse datasets (see Figure \ref{fig:teaser} and Figure \ref{fig:stylegan_main} (b)). Our method is capable of discovering several fine-grained directions, such as changing \textit{car type} or \textit{scenery} in LSUN Cars, changing \textit{breed} or adding \textit{fur} in LSUN Cats, adding \textit{windows} or \textit{turning on lights} in LSUN Bedrooms, adding \textit{tower} details to churches in LSUN Church, and adding \textit{riders} in LSUN Horse datasets. 

\paragraph{Comparison with other methods.} We compare how the directions found on FFHQ differ across methods. Figure \ref{fig:stylegan_main} (a) shows the visual comparison between several directions found in common by all methods, including the directions \textit{Smile, Lipstick, Elderly, Curly Hair }, and \textit{Young}. As can be seen from the figures, all methods perform similarly and are able to manipulate the images towards the desired attributes. Figure \ref{fig:stylegan_small} (a) illustrates various directions including \textit{Race, Eyeglass}, and \textit{Pose}. Figure \ref{fig:stylegan_small} (b) illustrates three different \textit{smile} directions discovered by our non-linear model.

\paragraph{Re-scoring analysis.} To understand how our method compares to the competitors in quantitative terms, we performed a re-scoring analysis \cite{shen2020closed} using attribute predictors to understand whether the manipulations changed the images towards the desired attributes. We used attribute predictors released by StyleGAN2 for the directions \textit{Smile} and \textit{Lipstick}, as these are the only two attributes for which a predictor is available, and which are simultaneously found in common by all methods. For the \textit{Age} direction, we used an off-the-shelf age predictor \cite{serengil2020lightface}. Table \ref{tab:stylegan2} shows our results for the re-scoring analysis for three attributes: \textit{Age}, \textit{Lipstick}, and \textit{Smile} for negative (labelled $\downarrow$) and positive (labelled $\uparrow$) directions. To see how the scores change, 500 images are  randomly generated for each property. The average score of the predictors for \textit{Lipstick} property is 0.43 $\pm 0.5$, the average score for \textit{Age} is 0.27 $\pm 0.1$, and the average score for \textit{Smile} is 0.74 $\pm 0.43$. Moving the latent codes in the negative direction, we find that our method, as well as the Ganspace and SeFa methods, decrease scores in similar ranges on the \textit{Age} and \textit{Lipstick} properties. We find that both our method and Ganspace are able to significantly reduce the score (from $0.74$ to $0.11$ and $0.05$, respectively) when moving towards the negative direction for the \textit{Smile} property, while SeFa reduces the score from $0.74$ to $0.50$ on average. When we move the latent codes in the positive direction, all methods achieve comparable results.

\section{Limitations}
\label{sec:limitations}
Our method uses a pre-trained GAN model as input, so it is limited to manipulating GAN-generated images only. However, it can be extended to real images by using GAN inversion methods \cite{zhu2020domain} by encoding the real images into the latent space. Like any image synthesis tool, our method also poses similar concerns and dangers in terms of misuse, as it can be applied to images of people or faces for malicious purposes, as discussed in \cite{korshunov2018deepfakes}. 
 
\section{Conclusion}
\label{sec:conclusion}
In this study, we propose a framework that uses contrastive learning to learn unsupervised directions. Instead of discovering fixed directions such as \cite{voynov2020unsupervised, harkonen2020ganspace, shen2020closed}, our method can discover non-linear directions in pre-trained StyleGAN2 and BigGAN models and leads to multiple distinct and semantically meaningful directions that are highly transferable. We demonstrate the effectiveness of our approach on a variety of models and datasets, and compare it to state-of-the-art unsupervised methods. We make our implementation available at \url{https://github.com/catlab-team/latentclr}.

\paragraph{Acknowledgments}
This publication has been produced benefiting from the 2232 International Fellowship for Outstanding Researchers Program of TUBITAK (Project No: 118c321). We also acknowledge the support of NVIDIA Corporation through the donation of the TITAN X GPU and GCP research credits from Google. We thank to Irem Simsar for proof-reading our paper. 

{\small
\bibliographystyle{ieee_fullname}
\bibliography{egbib}

\begin{thebibliography}{10}\itemsep=-1pt

\bibitem{abdal2021styleflow}
Rameen Abdal, Peihao Zhu, Niloy~J Mitra, and Peter Wonka.
\newblock Styleflow: Attribute-conditioned exploration of stylegan-generated
  images using conditional continuous normalizing flows.
\newblock {\em ACM Transactions on Graphics (TOG)}, 40(3):1--21, 2021.

\bibitem{BigGAN}
Andrew Brock, Jeff Donahue, and Karen Simonyan.
\newblock Large scale {GAN} training for high fidelity natural image synthesis.
\newblock {\em CoRR}, abs/1809.11096, 2018.

\bibitem{chen2020simple}
Ting Chen, Simon Kornblith, Mohammad Norouzi, and Geoffrey Hinton.
\newblock A simple framework for contrastive learning of visual
  representations.
\newblock In {\em International conference on machine learning}, pages
  1597--1607. PMLR, 2020.

\bibitem{chen2016infogan}
Xi Chen, Yan Duan, Rein Houthooft, John Schulman, Ilya Sutskever, and Pieter
  Abbeel.
\newblock Infogan: Interpretable representation learning by information
  maximizing generative adversarial nets.
\newblock {\em arXiv preprint arXiv:1606.03657}, 2016.

\bibitem{cherepkov2021navigating}
Anton Cherepkov, Andrey Voynov, and Artem Babenko.
\newblock Navigating the gan parameter space for semantic image editing.
\newblock In {\em Proceedings of the IEEE/CVF Conference on Computer Vision and
  Pattern Recognition}, pages 3671--3680, 2021.

\bibitem{goetschalckx2019ganalyze}
Lore Goetschalckx, Alex Andonian, Aude Oliva, and Phillip Isola.
\newblock Ganalyze: Toward visual definitions of cognitive image properties.
\newblock In {\em Proceedings of the IEEE/CVF International Conference on
  Computer Vision}, pages 5744--5753, 2019.

\bibitem{NIPS2014_5423}
Ian Goodfellow, Jean Pouget-Abadie, Mehdi Mirza, Bing Xu, David Warde-Farley,
  Sherjil Ozair, Aaron Courville, and Yoshua Bengio.
\newblock Generative adversarial nets.
\newblock In Z. Ghahramani, M. Welling, C. Cortes, N.~D. Lawrence, and K.~Q.
  Weinberger, editors, {\em Advances in Neural Information Processing Systems
  27}, pages 2672--2680. Curran Associates, Inc., 2014.

\bibitem{hadsell2006dimensionality}
Raia Hadsell, Sumit Chopra, and Yann LeCun.
\newblock Dimensionality reduction by learning an invariant mapping.
\newblock In {\em 2006 IEEE Computer Society Conference on Computer Vision and
  Pattern Recognition (CVPR'06)}, volume~2, pages 1735--1742. IEEE, 2006.

\bibitem{harkonen2020ganspace}
Erik H{\"a}rk{\"o}nen, Aaron Hertzmann, Jaakko Lehtinen, and Sylvain Paris.
\newblock Ganspace: Discovering interpretable gan controls.
\newblock {\em arXiv preprint arXiv:2004.02546}, 2020.

\bibitem{jahanian2019steerability}
Ali Jahanian, Lucy Chai, and Phillip Isola.
\newblock On the" steerability" of generative adversarial networks.
\newblock {\em arXiv preprint arXiv:1907.07171}, 2019.

\bibitem{StyleGAN}
Tero Karras, Samuli Laine, and Timo Aila.
\newblock A style-based generator architecture for generative adversarial
  networks.
\newblock {\em CoRR}, abs/1812.04948, 2018.

\bibitem{karras2020analyzing}
Tero Karras, Samuli Laine, Miika Aittala, Janne Hellsten, Jaakko Lehtinen, and
  Timo Aila.
\newblock Analyzing and improving the image quality of stylegan.
\newblock In {\em Proceedings of the IEEE/CVF Conference on Computer Vision and
  Pattern Recognition}, pages 8110--8119, 2020.

\bibitem{khrulkov2020disentangled}
Valentin Khrulkov, Leyla Mirvakhabova, Ivan Oseledets, and Artem Babenko.
\newblock On disentangled representations extracted from pretrained gans.
\newblock 2020.

\bibitem{korshunov2018deepfakes}
Pavel Korshunov and S{\'e}bastien Marcel.
\newblock Deepfakes: a new threat to face recognition? assessment and
  detection.
\newblock {\em arXiv preprint arXiv:1812.08685}, 2018.

\bibitem{li2019single}
Siyuan Li, Iago~Breno Araujo, Wenqi Ren, Zhangyang Wang, Eric~K. Tokuda,
  Roberto~Hirata Junior, Roberto Cesar-Junior, Jiawan Zhang, Xiaojie Guo, and
  Xiaochun Cao.
\newblock Single image deraining: A comprehensive benchmark analysis, 2019.

\bibitem{lin2020infogan}
Zinan Lin, Kiran Thekumparampil, Giulia Fanti, and Sewoong Oh.
\newblock Infogan-cr and modelcentrality: Self-supervised model training and
  selection for disentangling gans.
\newblock In {\em International Conference on Machine Learning}, pages
  6127--6139. PMLR, 2020.

\bibitem{mirza2014conditional}
Mehdi Mirza and Simon Osindero.
\newblock Conditional generative adversarial nets.
\newblock {\em arXiv preprint arXiv:1411.1784}, 2014.

\bibitem{nitzan2020face}
Yotam Nitzan, Amit Bermano, Yangyan Li, and Daniel Cohen-Or.
\newblock Face identity disentanglement via latent space mapping.
\newblock {\em arXiv preprint arXiv:2005.07728}, 2020.

\bibitem{noble2006support}
William~S Noble.
\newblock What is a support vector machine?
\newblock {\em Nature biotechnology}, 24(12):1565--1567, 2006.

\bibitem{oord2018representation}
Aaron van~den Oord, Yazhe Li, and Oriol Vinyals.
\newblock Representation learning with contrastive predictive coding.
\newblock {\em arXiv preprint arXiv:1807.03748}, 2018.

\bibitem{plumerault2020controlling}
Antoine Plumerault, Herv{\'e}~Le Borgne, and C{\'e}line Hudelot.
\newblock Controlling generative models with continuous factors of variations.
\newblock {\em arXiv preprint arXiv:2001.10238}, 2020.

\bibitem{radford2015unsupervised}
Alec Radford, Luke Metz, and Soumith Chintala.
\newblock Unsupervised representation learning with deep convolutional
  generative adversarial networks.
\newblock {\em arXiv preprint arXiv:1511.06434}, 2015.

\bibitem{ren2021generative}
Xuanchi Ren, Tao Yang, Yuwang Wang, and Wenjun Zeng.
\newblock Do generative models know disentanglement? contrastive learning is
  all you need.
\newblock {\em arXiv preprint arXiv:2102.10543}, 2021.

\bibitem{ImageNet}
Olga Russakovsky, Jia Deng, Hao Su, Jonathan Krause, Sanjeev Satheesh, Sean Ma,
  Zhiheng Huang, Andrej Karpathy, Aditya Khosla, Michael Bernstein,
  Alexander~C. Berg, and Li Fei-Fei.
\newblock {ImageNet Large Scale Visual Recognition Challenge}.
\newblock {\em International Journal of Computer Vision (IJCV)},
  115(3):211--252, 2015.

\bibitem{serengil2020lightface}
Sefik~Ilkin Serengil and Alper Ozpinar.
\newblock Lightface: A hybrid deep face recognition framework.
\newblock In {\em 2020 Innovations in Intelligent Systems and Applications
  Conference (ASYU)}, pages 1--5. IEEE, 2020.

\bibitem{shen2020interfacegan}
Yujun Shen, Ceyuan Yang, Xiaoou Tang, and Bolei Zhou.
\newblock Interfacegan: Interpreting the disentangled face representation
  learned by gans.
\newblock {\em IEEE Transactions on Pattern Analysis and Machine Intelligence},
  2020.

\bibitem{shen2020closed}
Yujun Shen and Bolei Zhou.
\newblock Closed-form factorization of latent semantics in gans.
\newblock {\em arXiv preprint arXiv:2007.06600}, 2020.

\bibitem{sohn2016improved}
Kihyuk Sohn.
\newblock Improved deep metric learning with multi-class n-pair loss objective.
\newblock In {\em Proceedings of the 30th International Conference on Neural
  Information Processing Systems}, pages 1857--1865, 2016.

\bibitem{spingarn2020gan}
Nurit Spingarn-Eliezer, Ron Banner, and Tomer Michaeli.
\newblock Gan" steerability" without optimization.
\newblock {\em arXiv preprint arXiv:2012.05328}, 2020.

\bibitem{Sun_2020}
Wanjie Sun and Zhenzhong Chen.
\newblock Learned image downscaling for upscaling using content adaptive
  resampler.
\newblock {\em IEEE Transactions on Image Processing}, 29:4027–4040, 2020.

\bibitem{tian2019contrastive}
Yonglong Tian, Dilip Krishnan, and Phillip Isola.
\newblock Contrastive multiview coding.
\newblock {\em arXiv preprint arXiv:1906.05849}, 2019.

\bibitem{tov2021designing}
Omer Tov, Yuval Alaluf, Yotam Nitzan, Or Patashnik, and Daniel Cohen-Or.
\newblock Designing an encoder for stylegan image manipulation.
\newblock {\em ACM Transactions on Graphics (TOG)}, 40(4):1--14, 2021.

\bibitem{voynov2020unsupervised}
Andrey Voynov and Artem Babenko.
\newblock Unsupervised discovery of interpretable directions in the gan latent
  space.
\newblock In {\em International Conference on Machine Learning}, pages
  9786--9796. PMLR, 2020.

\bibitem{wang2019spatial}
Tianyu Wang, Xin Yang, Ke Xu, Shaozhe Chen, Qiang Zhang, and Rynson Lau.
\newblock Spatial attentive single-image deraining with a high quality real
  rain dataset, 2019.

\bibitem{wang2017highresolution}
Ting-Chun Wang, Ming-Yu Liu, Jun-Yan Zhu, Andrew Tao, Jan Kautz, and Bryan
  Catanzaro.
\newblock High-resolution image synthesis and semantic manipulation with
  conditional gans, 2017.

\bibitem{wold1987principal}
Svante Wold, Kim Esbensen, and Paul Geladi.
\newblock Principal component analysis.
\newblock {\em Chemometrics and intelligent laboratory systems}, 2(1-3):37--52,
  1987.

\bibitem{yu2015lsun}
Fisher Yu, Ari Seff, Yinda Zhang, Shuran Song, Thomas Funkhouser, and Jianxiong
  Xiao.
\newblock Lsun: Construction of a large-scale image dataset using deep learning
  with humans in the loop.
\newblock {\em arXiv preprint arXiv:1506.03365}, 2015.

\bibitem{DBLP:journals/corr/abs-1710-10916}
Han Zhang, Tao Xu, Hongsheng Li, Shaoting Zhang, Xiaogang Wang, Xiaolei Huang,
  and Dimitris~N. Metaxas.
\newblock Stackgan++: Realistic image synthesis with stacked generative
  adversarial networks.
\newblock {\em CoRR}, abs/1710.10916, 2017.

\bibitem{CycleGAN}
Jun{-}Yan Zhu, Taesung Park, Phillip Isola, and Alexei~A. Efros.
\newblock Unpaired image-to-image translation using cycle-consistent
  adversarial networks.
\newblock {\em CoRR}, abs/1703.10593, 2017.

\bibitem{zhu2020domain}
Jiapeng Zhu, Yujun Shen, Deli Zhao, and Bolei Zhou.
\newblock In-domain gan inversion for real image editing.
\newblock In {\em European Conference on Computer Vision}, pages 592--608.
  Springer, 2020.

\end{thebibliography}
}

\end{document}